\documentclass[10pt,twocolumn,letterpaper]{article}

\usepackage[pagenumbers]{cvpr} 

\usepackage{graphicx}
\usepackage{amsmath}
\usepackage{amssymb}
\usepackage{booktabs}

\usepackage[pagebackref,breaklinks,colorlinks]{hyperref}

\usepackage[capitalize]{cleveref}
\crefname{section}{Sec.}{Secs.}
\Crefname{section}{Section}{Sections}
\Crefname{table}{Table}{Tables}
\crefname{table}{Tab.}{Tabs.}

\def\cvprPaperID{5192} 
\def\confName{CVPR}
\def\confYear{2022}

\def\TODO#1{\textcolor{red}{{\bf [TODO:}~#1{\bf ]}}~}
\def\NOTE#1{\textcolor{blue}{{\bf [NOTE:}~#1{\bf ]}}~}
\def\CHK#1 {\textcolor{magenta}{{\bf [CHK:}~#1{\bf ]}}~}
\def\ADD#1 {\textcolor{cyan}{{\bf [To add:}~#1{\bf ]}}~}
\def\etc{\emph{etc.} }
\def\ie{\emph{i.e.,} }
\def\eg{\emph{e.g.,} }
\def\etal{\emph{et al.} }
\def\Vec#1{{\boldsymbol{#1}}}
\def\Mat#1{{\boldsymbol{#1}}}
\usepackage{multirow}
\usepackage[flushleft]{threeparttable}
\usepackage{tabularx}
\usepackage{booktabs}

\begin{document}

\title{ Knowledge-enhanced Multi-perspective \\ Video Representation Learning for Scene Recognition }


\author{\textit{Xuzheng Yu}$^{1,2}$, \textit{Chen Jiang}$^2$, \textit{Wei Zhang}$^2$, \textit{Tian Gan}$^1$ \\
\textit{Linlin Chao}$^{2}$, \textit{Jianan Zhao}$^2$, \textit{Yuan Cheng}$^2$, \textit{Qingpei Guo}$^2$, \textit{Wei Chu}$^2$\\ $^1$Shandong University\\$^2$Ant Group
}

\maketitle

\begin{abstract}
  With the explosive growth of video data in real-world applications, a comprehensive representation of videos becomes increasingly important. In this paper, we address the problem of video scene recognition, whose goal is to learn a high-level video representation to classify scenes in videos.
  Due to the diversity and complexity of video contents in realistic scenarios, this task remains a challenge. 
  Most existing works identify scenes for videos only from visual or textual information in a temporal perspective, ignoring the valuable information hidden in single frames, while several earlier studies only recognize scenes for separate images in a non-temporal perspective.
  We argue that these two perspectives are both meaningful for this task and complementary to each other, meanwhile, external introduced knowledge can also promote the comprehension of videos.
  We propose a novel two-stream framework to model video representations from multiple perspectives, i.e. temporal and non-temporal perspectives, and integrate the two perspectives in an end-to-end manner by self-distillation. 
  Besides, we design a knowledge-enhanced feature fusion and label prediction method that contributes to naturally introducing knowledge into the task of video scene recognition.
  Experiments conducted on a real-world dataset demonstrate the effectiveness of our proposed method.
\end{abstract}

\section{INTRODUCTION}
    \begin{figure}[ht]
	\centering
	\includegraphics[width=0.47\textwidth]{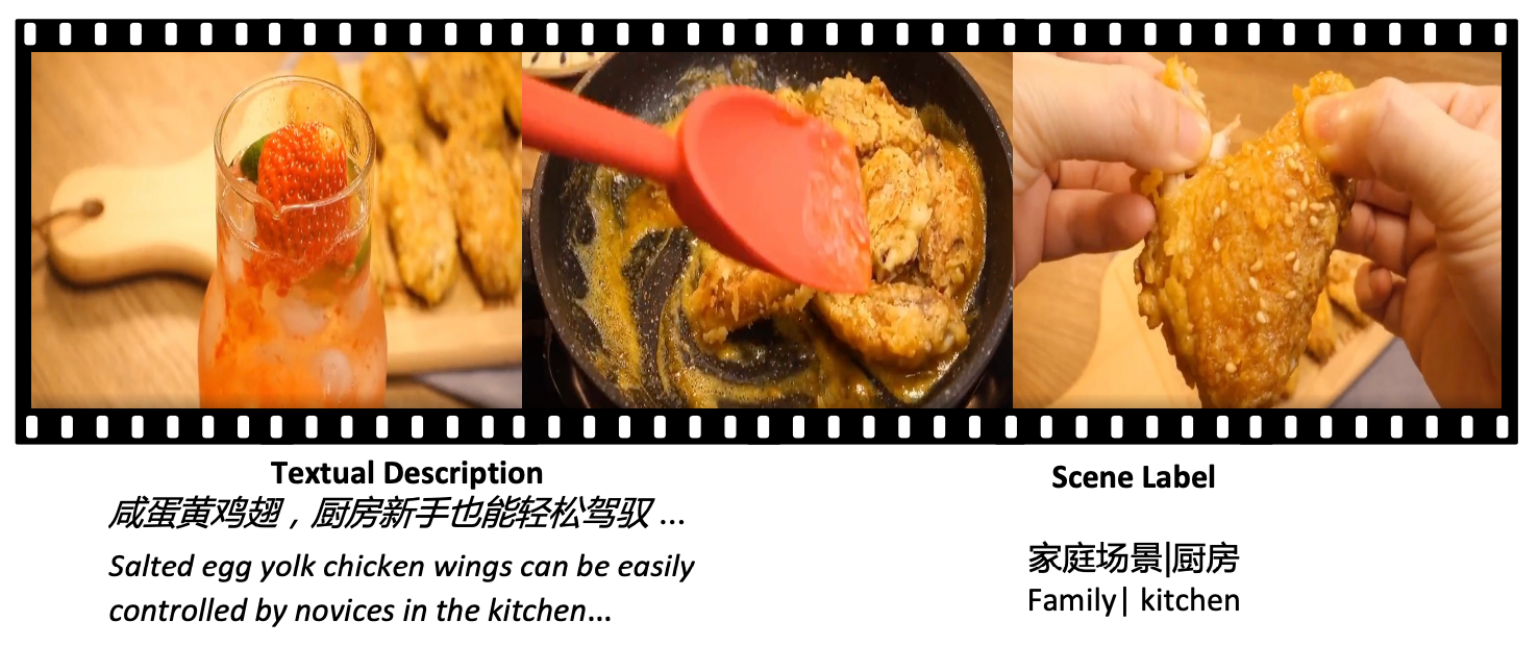} 
	\caption{An illustration of the task of scene recognition.
	}
	\label{fig_concept}
\end{figure}
    With the explosive growth of video data in real-world applications, analysis and reasoning on video content become increasingly important. One important branch is scene recognition from videos, which is to identify scene labels based on the content of videos.
    
    Video understanding itself is complex, including the diversity and redundancy of video contents, and the inherent gap among video multi-modal information.
    Therefore, the first intuitive challenge is how to effectively and comprehensively understand the contents of videos for this task. 
    Based on a deep insight of the characteristics of video scene recognition, it is notable that there exists multi-perspective information, including global vs local, temporal vs non-temporal, visual vs textural, etc. However, the diversity and discreteness of multiple information make it difficult to distinguish between useful and useless information. Meanwhile, we note that knowledge enhancement has been proven effective in many tasks, largely due to its modeling distinctiveness that contributes to constructing and reweighing a relation among multi-perspective information. Inspired by this,  we hope to introduce the knowledge enhancement into the task of video scene recognition to achieve an improvement.
   However, due to the conflict between the domain specificity and generality of knowledge enhancement, and the lack of general and excellent ways of fusing multi-perspective information with knowledge in the field of video understanding, how to leverage the external knowledge in our model is the second challenge.
    Moreover, knowledge-enhanced models are usually large-scale with low computational efficiency, and we expect to balance the efficiency and the additional time-consuming of introducing knowledge. Hence, it brings the third challenge that how to make the model as efficient as possible.
    
    In recent years, there has been an increasing amount of literature on video understanding \cite{ACMMM_HGLTM_2019, CVPR_STGK_2020, ACMMM_POET_2020}. These works are usually based on spatio-temporal relationship modeling, and build models for video from multiple perspectives. However, due to paying too much attention to the temporal information of videos, these works may ignore or lose the non-temporal information to varying degrees. In addition, several studies have revealed that external knowledge is beneficial for obtaining better performance \cite{AAAI_KBERT_2020, WWW_ESRASK_2017, ACL_EDNR_2018}. Specifically, these studies focus on linking linguistic tokens to entities in knowledge graphs, and enhancing reasoning based on the neighborhood of entities, while lacking attention to vision due to scarce general visual entity linking. Meanwhile, numerous studies have also focused on the reasoning efficiency \cite{CVPR_STGK_2020, CORR_DKNN_2015, ICLR_UDPI_2016}. These studies employ various methods to distill models, which give us great inspiration.

    In this paper, we propose a novel two-stream framework to learn video representations from multiple perspectives. Additionally, we design a knowledge-enhanced feature fusion method to integrate the multi-perspective information, which makes it natural to introduce knowledge into label recognition tasks. Finally, the application of knowledge distillation is employed to fuse the multiple representations, hoping to achieve high performance with low computational costs.
    
    Our main contributions are as follows:
    \begin{itemize}
        \item We propose a novel two-stream framework to model videos from temporal and non-temporal perspectives, and integrate the two perspectives through knowledge distillation in order to better comprehend videos while maintaining efficiency.
        
        \item We design a knowledge-enhanced feature fusion method to leverage external knowledge to better integrate features non-temporally, and introduce knowledge into the scene label prediction task naturally while additionally gaining label scalability as a by-product.
        
        \item We quantitatively evaluated our model on a real-world dataset. Experimental results demonstrate the effectiveness of our model.
    \end{itemize}

   
    The rest of this paper is structured as follows. In Section 2, we briefly review the related literature. In Section 3, we detail our proposed model, followed by experimental results and analyses in Section 4. We finally conclude the work in Section 5.

\section{RELATED WORK}

In this section, we mainly review the studies that are most related to our work, 
including video representation,
scene recognition,
and knowledge-enhanced learning.

\subsection{Video Representation}
    Obtaining video representation is essential and indispensable for video analytics, 
    while for obtaining video representation,
    spatio-temporal modeling of videos is an important step \cite{IEEE_SMVS_2019, CVPR_LSTR_2019, ICCV_LSTR_2017}.
    There are many attempts have been made to model videos from spatial and temporal perspectives \cite{ACMMM_HGLTM_2019, CVPR_STGK_2020, CVPR_ORGTRL_2020, ACMMM_LSCTA_2020}.
    Hu \etal \cite{ACMMM_HGLTM_2019} proposed a hierarchical temporal method to construct the temporal structure at frame-level and object-level successively, and extract pivotal information effectively from global to local, which improves the model capacity of recognizing fine-grained objects and actions. 
    Pan \etal \cite{CVPR_STGK_2020} proposed a novel spatio-temporal graph network to explicitly exploit the spatio-temporal object interaction, which is crucial for video understanding and description.
    Besides, Shi \etal \cite{ACMMM_LSCTA_2020} proposed to learn the semantic concepts explicitly and design a temporal alignment mechanism to better align the video and transcript.
    These studies give us a lot of inspiration, 
    however,
    they are too focused on the impact of temporal information on video comprehension due to task constraints, while ignoring non-temporal information to some extent.

\subsection{Scene Recognition}
    Scene recognition is a task to develop robust and reliable models for the automatic recognition of what scenes are described by visual information \cite{HAIS_DLSR_2020}. Early research on scene recognition mainly focus on separate images \cite{TRANS_PLACES365_2018}, while scholars' attention naturally turn towards scene recognition from videos \cite{TRANS_HCMFL_2019, TRANS_HETFN_2021}.
    Zhou \etal \cite{TRANS_PLACES365_2018} described the Places Database, and provided scene recognition convolutional neural networks as baselines.
    Jiang \etal \cite{TRANS_HCMFL_2019} proposed a novel framework, which jointly utilized multi-platform data, object-scene deep features and the hierarchical venue structure prior for scene category prediction from videos.
    Zhang \etal \cite{TRANS_HETFN_2021} proposed a Hybrid-Attention Enhanced Two-Stream Fusion Network for the task of video scene label prediction, and develops a novel Global-Local Attention Module, which can be inserted into neural networks to generate enhanced visual features from video content.
    Most recent studies identify scenes only from visual or textual information in a temporal perspective, ignoring the valuable information hidden in single frames, while several earlier studies only recognize scenes for separate images in non-temporal perspective. In this paper, we argue these two perspectives are both meaningful for this task and complementary to each other.
\subsection{Knowledge-enhanced Learning}

    Knowledge-enhanced learning is increasingly attracting more attention from researchers in recent years. 
    To make models better mine the hidden information in data, 
    many scholars tried to introduce different types of additional information into their methods \cite{AAAI_KBERT_2020, WWW_ESRASK_2017, ACMMM_POET_2020, ACMMM_KPMPE_2021, CORR_DKNN_2015, ICLR_UDPI_2016, CVPR_STGK_2020}. 
    This additional information is considered as knowledge, and is usually related to knowledge graphs \cite{AAAI_KBERT_2020, WWW_ESRASK_2017}, transferred knowledge \cite{CORR_DKNN_2015, ICLR_UDPI_2016, CVPR_STGK_2020},
    and specifically defined knowledge \cite{ACMMM_POET_2020, ACMMM_KPMPE_2021}.

    When it comes to knowledge, scholars often refer to knowledge graphs and the embedding representations of the nodes and edges in knowledge graphs. 
    Several studies \cite{AAAI_KBERT_2020, WWW_ESRASK_2017, SIGIR_WEDR_2017, ACL_EDNR_2018, AAAI_CONCEPTNET_2017} have been carried out on knowledge graphs and verified that the pretrained embeddings from knowledge graphs are helpful to reasoning.
    Liu \etal \cite{AAAI_KBERT_2020} proposed a knowledge-enabled language representation model with knowledge graphs, 
    in which triples are injected into the sentences as domain knowledge. Xiong \etal \cite{WWW_ESRASK_2017} introduced a novel method to represent queries and documents in the entity space, and rank documents based on their semantic relatedness to queries. These studies fully explore the node embeddings and edges in knowledge graphs, 
    while they mainly focus on text modality.
    And in our work, we leverage the pretrained knowledge embeddings to guide cross-modality fusion instead of single-modal reasoning. 

    Compared with knowledge graphs focusing on mining the valuable information hidden in nodes and edges in themselves, transferred knowledge focus on the transmission and sharing of known valuable information (\eg distributions).
    The transferred knowledge is usually related to the pretrained models \cite{NAACL_BERT_2019, CVPR_RESNET_2016}, 
    especially pretrained language models \cite{NAACL_BERT_2019}, 
    and the transferred distributions are common in the task of knowledge distillation \cite{CVPR_STGK_2020, CVPR_ORGTRL_2020, CORR_DKNN_2015, ICLR_UDPI_2016}.
    Pan \etal \cite{CVPR_STGK_2020} proposed a novel spatio-temporal graph network and designed a two-branch framework with an object-aware knowledge distillation mechanism.
    Zhang \etal \cite{CVPR_ORGTRL_2020} proposed a novel teacher-recommended learning method that introduces external language model to guide the main model to learn abundant linguistic knowledge. 
    In these methods, 
    for better reasoning, knowledge is transferred between multiple modules that perform the same task in different ways, 
    and this idea is 
    adopted in our work.

    The specifically defined knowledge in certain scenarios also appears in numerous recent studies \cite{ACMMM_POET_2020, ACMMM_KPMPE_2021}.
    Zhang \etal \cite{ACMMM_POET_2020} proposed to narrate the user-preferred product characteristics depicted in user-generated product videos, and proposed a novel framework to perform knowledge-enhanced spatio-temporal inference on product-oriented video graphs.
    Zhu \etal \cite{ACMMM_KPMPE_2021} introduced knowledge modality in multi-modal pretraining to correct the noise and supplement the missing image and text modalities. 
    The aforementioned methods introduce several types of additional associated information, and leverage the specifically defined knowledge to guide the learning of models. However, the introduced knowledge also brings barriers to these methods, 
    because the knowledge and methods usually target specific datasets,
    and are not easy to be extended to other tasks. 
    In our work, 
    though we similarly introduce additional specifically defined information (\ie keywords) as knowledge, 
    yet the introduced keywords are easily obtained from text descriptions, 
    making our method extendable and adjustable.

\section{OUR PROPOSED METHOD}
\begin{figure*}[ht]
	\centering
	\includegraphics[width=1\textwidth]{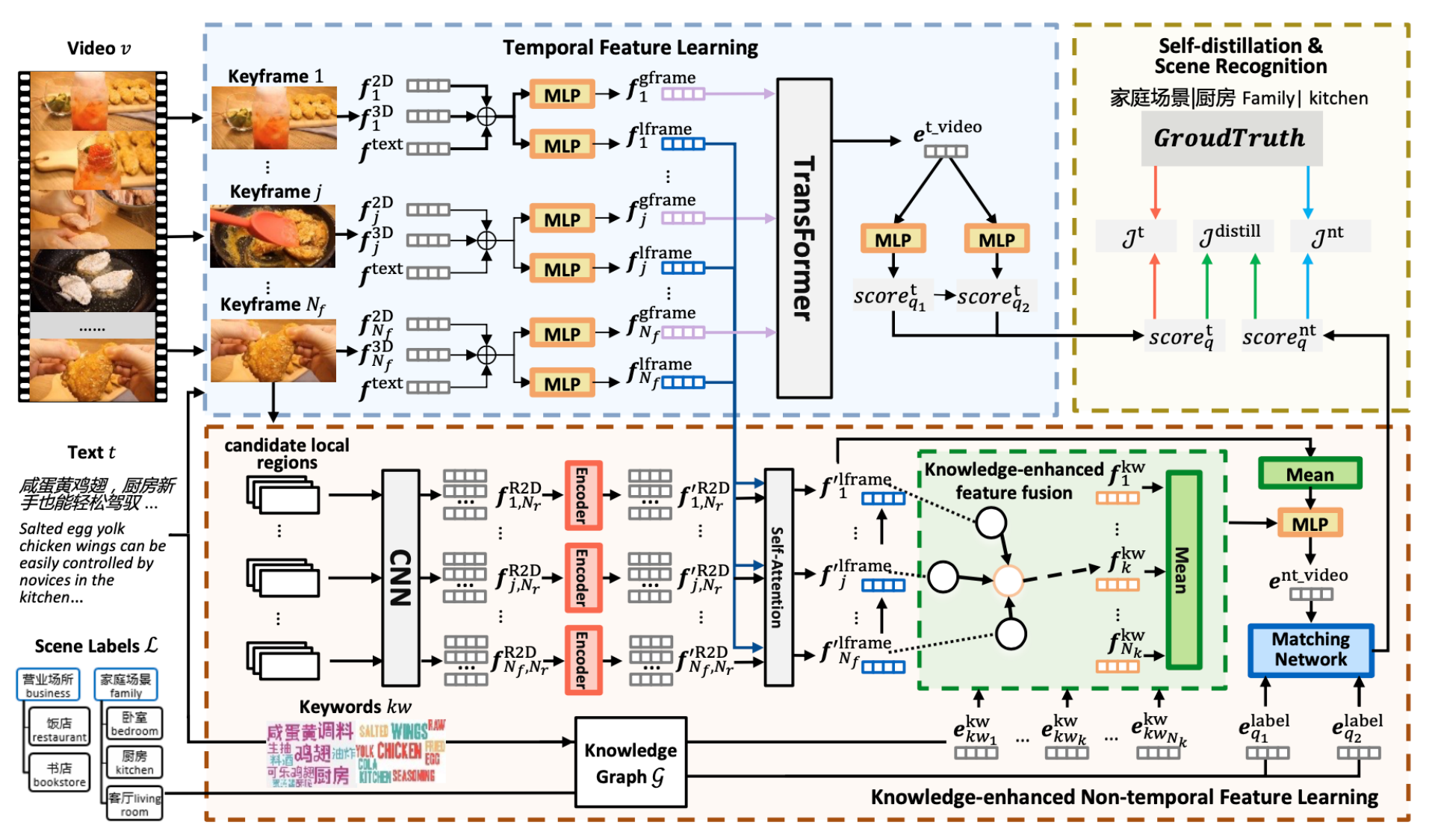} 
	\caption{An overview of our proposed framework for scene recognition using knowledge-enhanced multi-perspective video representation learning.
	}
	\label{fig_overview}
\end{figure*}
\subsection{Overview}
    \subsubsection{Problem setting.} 
        Before describing our method, we briefly introduce the problem setting first. 
        Formally, let $\mathcal{V}$, 
        $\mathcal{L}$, 
        and $\mathcal{G}$ denote the set of videos, 
        the set of scene labels, 
        and the external knowledge graph respectively.
        Each video $v \in \mathcal{V}$ contains textual descriptions $t$ and a sequence of RGB frames. 
        Scene labels $\mathcal{L}$ belong to a two-level scene hierarchy, and each video is associated with a set of paths $\mathcal{P}_{v}=\{ p_{1}, p_{2}, ..., p_{|\mathcal{P}_{v}|}\}$ on this hierarchy, where $p_{l}=\{ p^{1}_{l}, p^{2}_{l} \ | \ p^{1}_{l}, p^{2}_{l} \in \mathcal{L}\}$ for $p_{l}\in \mathcal{P}_{v}$, and $p^{1}_{l}$ is the parent scene label of $p^{2}_{l}$. 
        Our goal is to learn a video scene recognition model, which could recognize suitable scene labels $p$ based on the video frame sequences and the associated text descriptions.
    \subsubsection{Model overview.}
        Temporal information can well describe what happens in videos, while non-temporal information can well reflect several moments in videos. 
        We argue that these two perspectives are both meaningful for this task and complementary to each other.
        Meanwhile, external introduced knowledge can also promote the comprehension of videos. 
        In order to make full use of these kinds of information, we propose a novel 
        two-stream framework to model video representations from the two perspectives, i.e. temporal and non-temporal perspectives, 
        and integrate these two perspectives in an end-to-end manner by self-distillation. 
        Besides, in the non-temporal module, 
        we design a knowledge-enhanced feature fusion and label prediction
        method that contributes to naturally introducing knowledge into the task of video scene recognition.

\subsection{Temporal Feature Learning}
    The temporal feature learning module is used to model videos from a temporal perspective, 
    including temporal modeling and hierarchical multi-label prediction.
    \subsubsection{Temporal modeling.}
        
        In this module, we employ Multimodal Bitransformers \cite{NEURIPS_MMBT_2019} as the backbone, 
        and utilize Transformer \cite{NEURIPS_TRANSFORMER_2017} that is effective in various tasks in recent years as the feature encoder. 
        %
        Specifically, 
        for each video $v$, 
        we uniformly sample $N_{\rm{f}}$ frames as keyframes, 
        and utilized the pretrained ResNet \cite{CVPR_RESNET_2016} to extract the frame-level 2D feature
        $\Vec{f}^{\rm{2D}}_{j}$ of each keyframe, 
        where $j \in [1,N_{\rm{f}}]$. 
        We then collect $N_{\rm{c}}$ consecutive frames with each sampled keyframe as center, 
        and utilized the pretrained I3D \cite{CVPR_I3D_2017} to extract the frame-level 3D features $\Vec{f}^{\rm{3D}}_{j}$. 
        For the associated textual descriptions, 
        we utilize the pretrained BERT \cite{NAACL_BERT_2019} to extract the video-level textual features $\Vec{f}^{\rm{text}}$.
        
        After extracting the aforementioned features, 
        we concatenate them to obatin the frame-level global features $\Vec{f}^{ \rm{gframe}}_{j}$ and local features $\Vec{f}^{ \rm{lframe}}_{j}$ as follows: 
        \begin{gather}
        \Vec{e}^{\rm{con\_\tau}}_{j}=(\Vec{f}^{\rm{2D}}_{j}   \operatorname{\oplus} \Vec{f}^{\rm{3D}}_{j} \operatorname{\oplus} \Vec{f}^{\rm{text}}), \\
        \Vec{e}^{\rm{ref\_\tau}}_{j}=  \Mat{W}^{\rm{\tau}}_{2} \operatorname{\delta}( \Mat{W}^{\rm{\tau}}_{1}  \Vec{e}^{\rm{con\_\tau}}_{j}+ \Vec{b}^{\rm{\tau}}_{1} )+ \Vec{b}^{\rm{\tau}}_{2}, \\
        \Vec{f}^{\rm{\tau}}_{j}= Norm( Dropout(\Vec{e}^{\rm{ref\_\tau}}_{j}) + \Mat{W}^{\rm{\tau}}_{3}\Vec{e}^{\rm{con\_\tau}}_{j}),\\
        \tau \in \{ \rm{gframe}, \rm{lframe} \}, \nonumber
        \end{gather}
        where $\operatorname{\oplus}$ indicates concatenation, $\Vec{e}^{\rm{con}}_{j}$ and $\Vec{e}^{\rm{ref}}_{j}$ indicate the concatenated and the refined features respectively, $\operatorname{\delta}(\cdot)$ indicates the activation function, $\Mat{W}_{1}$, $\Mat{W}_{2} $, $\Vec{b}_{1}$ and $\Vec{b}_{2}$ denote the weight matrices and bias vectors in fully connected layers, $\Mat{W}_{3}$ denotes the residual transformation matrices, and $Norm$ refers to the operation of layer normalization.

        In order to better model the temporal aspects of video,
        we send the special tag $[CLS]$ and the frame-level global features $\Vec{f}^{ \rm{gframe}}_{j}$ into the self-attention layers of the Transformer encoder with position embeddings. After the encoding, the Transformer encoder generates a sequence of outputs, and we denote the generated output corresponding to the special tag $[CLS]$ as the video-level temporal features $\Vec{e}^{\rm{t\_video}}$ as follows:
        \begin{gather}
        \begin{split}
        &\ \ \ \ \ \ \ \ \ \ \ \ \ \Vec{O} = \left[\Vec{o}_{0}, \Vec{o}_{1}, ..., \Vec{o}_{N_{\rm{f}}}\right]\\
        &=\ \operatorname{Transformer}([[CLS]+\Vec{pos}_{0},\\
        &\Vec{f}^{ \rm{gframe}}_{1}+\Vec{pos}_{1}, ... , \Vec{f}^{ \rm{gframe}}_{N_f}+\Vec{pos}_{N_{\rm{f}}}]),
        \end{split}\\
        \Vec{e}^{\rm{t\_video}}=\Vec{o}_{0},
        \end{gather}
        where $\Vec{O}$ indicates the output sequence, $\Vec{pos}_{j} $ ( $j\in\left[0, N_{\rm{f}} \right]$ ) indicates the position embedding, and $N_{\rm{f}}$ denotes the number of sampled keyframes.
        
    \subsubsection{Hierarchical multi-label prediction.}
        After obtaining the video-level temporal features $\Vec{e}^{\rm{t\_video}}$, we employ multi-layer perceptrons to obtain the basic scores of labels ${score\rm{'}}^{\rm{t}}_{i}$ for the $i^{\rm{th}}$-level scene hierarchical layer. Besides, for each label $q_1$ in the $1^{st}$-level scene hierarchical layer, we make the refined scores ${score}^{t}_{q_1}$ of it the same as its basic score. And for each label $q_2$ in the $2^{rd}$-level layer, we obtain the refined scores ${score}^{\rm{t}}_{q_2}$ of it by summing the basic scores of itself and its parent label $\mu_{q_2}$ as follows:
        \begin{align}
        {score\rm{'}}^{\rm{t}}_{i}=\Mat{W}^{i}_{5} \operatorname{\delta}( \Mat{W}^{i}_{4}\Vec{e}^{\rm{t\_video}}+\Vec{b}^{i}_{4} )+ \Vec{b}^{i}_{5}, \\
        {score}^{\rm{t}}_{q_1}={score\rm{'}}^{\rm{t}}_{1,q_1}, \\
        {score}^{\rm{t}}_{q_2}={score\rm{'}}^{\rm{t}}_{1,\mu_{q_2}}+{score\rm{'}}^{\rm{t}}_{2,q_{2}},
        \end{align}
        where $score^{\rm{t}}_{i} \in \mathbb{R}^{|\mathcal{L}_{i}|}$ and $|\mathcal{L}_{i}|$ denote the scores and the number of the labels in the $i^{\rm{th}}$-level scene hierarchical layer respectively, $\mu_{\theta}$ denotes the parent label of $\theta$ in scene hierarchy, $\operatorname{\delta}(\cdot)$ indicates the activation function, $\Mat{W}^{i}_{4} \in \mathbb{R}^{d_{\rm{emb}}\times d_{\rm{emb}}}$ and $\Mat{W}^{i}_{5}\in \mathbb{R}^{|\mathcal{L}_{i}|\times d_{\rm{emb}}}$ denote weight matrices in fully connected layers, $\Vec{b}^{i}_{4} \in \mathbb{R}^{d_{\rm{emb}}}$ and $\Mat{b}^{i}_{5}\in \mathbb{R}^{|\mathcal{L}_{i}|}$ denote bias vectors, and $d_{emb}$ denotes the dimensions of the video-level temporal features.

        And then we employ the Multi-label Cross Entropy Loss \cite{SS_MCE_2020, CVPR_CL_2020} to calculate the loss for the $i^{th}$-level scene hierarchical layer as follows:
        \begin{gather}
        \begin{split}
        \mathcal{J}_{i}^{\rm{t}}=  \frac{1}{|\mathcal{V}|} \sum_{v\in\mathcal{V}} \{ log[1+\sum_{q^{-}_i \notin \mathcal{P}_v^i} exp(score^{\rm{t}}_{q^{-}_{i}})]\\
        +log[1+\sum_{q^{+}_{i} \in \mathcal{P}_{v}^{i}} exp(-score^{\rm{t}}_{q^{+}_{i}})] \},
        \end{split}
        \end{gather}
        where $|\mathcal{V}|$ denotes the number of videos, $q^{+}_{i}$ and $q^{-}_{i}$ denotes the positive and the negative scene labels in the $i^{th}$-level scene hierarchy respectively, and $\mathcal{P}_{v}^{i}$ denotes the annotated scene labels of video $v$ in the $i^{th}$-level scene hierarchy.
        
        Finally, the final temporal objective function can be formulated as follows:
        \begin{gather}
        \mathcal{J}^{\rm{t}}=\beta_{1}^{\rm{level}}\mathcal{J}_{1}^{\rm{t}}+\beta_{2}^{\rm{level}}\mathcal{J}_{2}^{\rm{t}},
        \end{gather}
        where $\beta_{1}^{\rm{level}}$ and $\beta_{2}^{rm{level}}$ are hyper-parameters as coefficients of $\mathcal{J}_{1}^{\rm{t}}$ and $\mathcal{J}_{2}^{\rm{t}}$.

\subsection{Knowledge-enhanced Non-temporal Feature Learning}
    The knowledge-enhanced non-temporal feature learning module is used to model videos from a knowledge-enhanced non-temporal perspective. We first obtain frame-level local features by fusing candidate local regions, and then introduce external knowledge to perform and enhance the video feature fusion and scene prediction. 
    
    \subsubsection{Frame-level local feature fusion.}
        
        In this module, we employ the pretrained Faster-RCNN \cite{NIPS_FRCNN_2015} to detect the candidate local regions, extract their 2D features with the pretrained ResNet, and denote the 2D feature of the $m$-th candidate regions of the $j$-th keyframe as $\Vec{f}^{\rm{R2D}}_{j,m}$. 
        
        Then we send all the candidate region features detected in each keyframe directly into the self-attention layers of the Transformer encoder. We employ this way to complete inner-frame reasoning among the candidate regions detected in the same keyframe, and obtain the enhanced feature ${\Vec{f}\rm{'}}^{\rm{R2D}}_{j,m} \in {\Vec{f}\rm{'}}^{\rm{R2D}}_{j}$ of each candidate region as follows:
        \begin{gather}
        {\Vec{f}\rm{'}}^{\rm{R2D}}_{j} = \operatorname{Transformer}( [\Vec{f}^{\rm{R2D}}_{j,1}, \Vec{f}^{\rm{R2D}}_{j,2}, ..., \Vec{f}^{\rm{R2D}}_{j,N_{\rm{r}}}] ),
        \end{gather}
        where $N_{\rm{r}}$ indicates the number of candidate regions extracted in the same keyframe.

        After that, we leverage the frame-level local features $\Vec{f}^{ \rm{lframe}}_{j}$  obtained in the previous section as queries, perform self-attention operations on the enhanced candidate region feature $\Vec{e}^{\rm{R2D}}_{j,m} \in \Vec{e}^{\rm{R2D}}_{j}$, and denote the obtained fused frame-level local features ${\Vec{f}\rm{'}}^{\rm{lframe}}_{j}$ as follows:
        \begin{gather}
        \begin{align}
        \Vec{Query}_{j} &= \Vec{f}^{ \rm{lframe}}_{j}\Mat{W}^{\rm{local}}_{\rm{Q}}, \\
        \Vec{Key}_{j,m} &= {\Vec{f}\rm{'}}^{\rm{R2D}}_{j,m}\Mat{W}^{\rm{local}}_{\rm{K}}, \\
        \Vec{Value}_{j,m} &= {\Vec{f}\rm{'}}^{\rm{R2D}}_{j,m}\Mat{W}^{\rm{local}}_{\rm{V}},
        \end{align}\\
        {\alpha}^{\rm{frame}}_{j,m}= \operatorname{softmax} ( \frac{\Vec{Query}_{j,m}\Vec{Key}_{j,m} }{\sqrt{d_{\rm{Key}}}} ), \\
        {\Vec{f}\rm{'}}^{\rm{lframe}}_{j}=\sum_{m \in [1,M]} {\alpha}^{\rm{frame}}_{j,m}\Vec{Value}_{j,m},
        \end{gather}
        where $\Mat{W}^{\rm{local}}_{\rm{Q}}$, $\Mat{W}^{\rm{local}}_{\rm{K}}$, $\Mat{W}^{\rm{local}}_{\rm{V}}$ indicate the weight matrices corresponding to the queries, keys and values of the self-attention module, and ${d}_{\rm{Key}}$ represents the dimensions of the key vectors.

    \subsubsection{Knowledge-enhanced video feature fusion.}
        In this module, we leverage the entity embeddings pretrained from knowledge graphs $\mathcal{G}$ as external knowledge, and denote $\Vec{G}_{\theta}$ as the pretrained embeddings corresponding to the token $\theta$. In addition, we also employ the embedding-based \cite{ICLR_WORD2VEC_2013, NAACL_BERT_2019} unsupervised keyword extraction algorithm to extract $N_{k}$ keywords ${kw}_{k} \in \mathcal{K}_{v} \subseteq \mathcal{G}$ from the associated text descriptions of video $v$.
        
        In terms of feature fusion, inspired by NetVLAD \cite{CVPR_NETVLAD_2016}, we design a knowledge-enhanced feature fusion method. Unlike NetVLAD, which directly learn a set of parameters for each cluster center and clusters features based on the measured distances, we employ shared parameter modules to generate the required parameters based on the features of the clustering target (\ie{the pretrained word embeddings of keywords}) as follows:
        \begin{gather}
        \operatorname{\varphi}_{\rm{w}} (\theta)=\Mat{W}^{\rm{w}}_{2} \operatorname{\delta}( \Mat{W}^{\rm{w}}_{1} \theta + \Vec{b}^{\rm{w}}_{1} )+ \Vec{b}^{\rm{w}}_{2}, \\
        \operatorname{\varphi}_{\rm{c}} (\theta)=\Mat{W}^{\rm{c}}_{2} \operatorname{\delta}( \Mat{W}^{\rm{c}}_{1} \theta + \Vec{b}^{\rm{c}}_{1} )+ \Vec{b}^{\rm{c}}_{2}, \\
        \operatorname{\varphi}_{\rm{z}} (\theta)=\Mat{W}^{\rm{z}}_{2} \operatorname{\delta}( \Mat{W}^{\rm{z}}_{1} \theta + \Vec{b}^{\rm{z}}_{1} )+ \Vec{b}^{\rm{z}}_{2},
        \end{gather}
        where $\Mat{W}^{\rm{\tau}}_{1}$, $\Mat{W}^{\rm{\tau}}_{2}$, $\Vec{b}^{\rm{\tau}}_{1}$ and $\Vec{b}^{\rm{\tau}}_{2}$ ( $\rm{\tau} \in\{ \rm{w}, \rm{c}, \rm{z} \}$ ) denote weight matrices and bias vectors in fully connected layers, $\operatorname{\varphi}_{\rm{w}} (\theta)$ and $\operatorname{\varphi}_{\rm{c}} (\theta)$ denote the functions which can generate parameters to measure distances, $\operatorname{\varphi}_{\rm{z}} (\theta)$ denotes the function which is usd to generate the representations of the cluster centers, and $\operatorname{\delta}(\cdot)$ indicates the activation function.

        Then we measure the similarity between different frame-level local features and keyword semantics based on the learned parameters, and weighted summing the difference between the frame-level local features and the obtained representations of the keyword cluster centers. Through this design, we can fuse the features of the sampled frames to varying degrees based on the semantics of specific keywords, and obtain the knowledge-enhanced keyword-level non-temporal feature $\Vec{f}^{\rm{kw}}_{k}$ as follows:
        \begin{gather}
        {\alpha}^{\rm{kw}}_{j,k}= \operatorname{softmax} (\operatorname{\varphi}_{\rm{w}}(\Mat{G}_{{kw}_{k}} ){\Vec{f}\rm{'}}^{\rm{lframe}}_{j}+\operatorname{\varphi}_{\rm{c}}(\Mat{G}_{{kw}_{k}})), \\
        \Vec{f}^{\rm{kw}}_{k}=\sum_{j \in [1,N_{\rm{f}}]} {\alpha}^{\rm{video}}_{j,k} ({\Vec{f}\rm{'}}^{\rm{lframe}}_{j}-\operatorname{\varphi}_{\rm{z}}(\Mat{G}_{{kw}_{k}} )),
        \end{gather}
        where ${kw}_{k}$ represents the $k$-th extracted keyword of video $v$, $N_{\rm{f}}$ denotes the number of the sampled keyframes of each video, and $G_{\theta}$ denotes the operation of obtaining the pretrained word embedding corresponding to the specific linguistic token ${\theta}$.
        
        After that, we utilize mean pooling to fuse multiple knowledge-enhanced keyword-level non-temporal features of the same video, and leverage the mean feature of the fused frame-level local features, to receive the knowledge-enhanced video-level non-temporal feature $\Vec{e}^{\rm{nt\_video}}$ for each video as follows:
        \begin{gather}
        {\Vec{e}\rm{'}}^{\rm{nt\_video}}=\frac{1}{|\mathcal{K}|} \sum_{k \in [1,|\mathcal{K}|]}\!\Vec{f}^{\rm{kw}}_{k}\!+\!\frac{1}{N_f}\!\sum_{j \in [1,N_f]} {\Vec{f}\rm{'}}^{\rm{lframe}}_{j}, \\
        \Vec{e}^{\rm{nt\_video}}=\Mat{W}^{\rm{nt}}  {\Vec{e}\rm{'}}^{\rm{nt\_video}}+ \Vec{b}^{\rm{nt}},
        \end{gather} 
        where $\Mat{W}^{\rm{nt}}$ and $\Vec{b}^{\rm{nt}}$ denote weight matrices and bias vectors in the fully connected layer, $|\mathcal{K}|$ denotes the number of the extracted keywords of videos, and $N_f$ denotes the number of the sampled keyframes for each video.
    
    \subsubsection{Knowledge-enhanced multi-label scene prediction}
        In order to make better use of knowledge for scene prediction, we design shared matching networks to calculate the basic matching scores ${score\rm{'}}^{\rm{nt}}_{v,q}$ between videos $v$ and the $i^{th}$-level scene label $q_i$, based on the knoweledge-enhanced video-level non-temporal featrue $\Vec{e}^{\rm{nt\_video}}_{v}$ and the scene label representations $\Vec{e}^{\rm{label}}_{q_i}$. After that, we obtain the refined scores $score^{\rm{nt\_video}}_{q_i}$ using approaches similar to those in the hierarchical multi-label prediction in the temporal module as follows:
        \begin{gather}
        {\Vec{e}\rm{'}}^{\rm{nt\_video}}_{v,i}=\Mat{W}^{i}_{7} \operatorname{\delta}( \Mat{W}^{i}_{6} \Vec{e}^{\rm{nt\_video}}_{v}+\Vec{b}^{i}_{6} )+ \Vec{b}^{i}_{7}, \\
        {\Vec{e}\rm{'}}^{\rm{label}}_{q_i}=\Mat{W}^{i}_{9} \operatorname{\delta}( \Mat{W}^{i}_{8} \Vec{e}^{\rm{label}}_{q_i}  + \Vec{b}^{i}_{8} )+ \Vec{b}^{i}_{9}, \\
        {score\rm{'}}^{\rm{nt}}_{v,q_i}={\Vec{e}\rm{'}}^{\rm{nt\_video}}_{v,i} \circ {\Vec{e}\rm{'}}^{\rm{label}}_{q_i}, \\
        {score}^{\rm{nt}}_{v,q_1}={score\rm{'}}^{\rm{nt}}_{v,q_1}, \\
        {score}^{\rm{nt}}_{v,q_2}={score\rm{'}}^{\rm{nt}}_{v,\mu_{q_2}}+{score\rm{'}}^{\rm{nt}}_{v,q_{2}},
        \end{gather}
        where $\Mat{W}^{i}_{\tau}$ and $\Vec{b}^{i}_{\tau}$ ($i \in \{1, 2\}$, $\tau \in\{ 6, 7, 8, 9 \}$ ) denote weight matrices and bias vectors in fully connected layers, $\operatorname{\delta}(\cdot)$ indicates the activation function, $\circ$ represents the inner product operation, and $\mu_{\theta}$ denotes the parent label of $\theta$ in scene hierarchy.
        
        For those scene labels whose corresponding entity can be found in the knowledge graph $\mathcal{G}$, we directly utilize their corresponding pretrained entity embeddings $G_{q}$ as the scene representations $e^{\rm{label}}_{q}$. And for the remaining unmatched scene labels, we initialize their embeddings randomly and update them when training the network. In this way, we also receive an additional benefit, which is the scalability of labels. Specifically, the scalability of labels means that, when we need to predict an unseen label that is not in the original label list but is an entity in knowledge graphs, we do not need to retrain the model, and can perform inferences directly.

        Similar to the temporal module, we also employ the Multi-label Cross Entropy Loss to calculate the loss $\mathcal{J}^{\rm{nt}}_{i}$ for each scene hierarchical layer, and the final non-temporal objective function can be formulated as follows:
        \begin{gather}
        \mathcal{J}^{\rm{nt}}=\beta_{1}^{\rm{level}}\mathcal{J}_{1}^{\rm{nt}}+\beta_{2}^{\rm{level}}\mathcal{J}_{2}^{\rm{nt}},
        \end{gather}
        where $\beta_{1}^{\rm{level}}$ and $\beta_{2}^{\rm{level}}$ are hyper-parameters as coefficients of $\mathcal{J}_{1}^{\rm{nt}}$ and $\mathcal{J}_{2}^{\rm{nt}}$.

\subsection{Self-distillation and Scene Recognition}
    In the aforementioned sections, we model videos from two perspectives (\ie{the temporal perspective and the knowledge-enhanced non-temporal perspective}), and obtain scene label scores from both perspectives.

    To enable the two modules (\ie{temporal module and the knowledge-enhanced non-temporal module}) to learn information separately, and integrate the learned information with each other, we employ $Euclidean distance$ to measure the difference between the two groups (\ie{the temporal perspective and the knowledge-enhanced non-temporal perspective}) of scene label scores in each scene hierarchical layer. We obtain the distillation loss $\mathcal{J}^{\rm{distill}}$ according to the obtained Euclidean distances $\mathcal{J}_{i}^{\rm{distill}}$ for the $i^{th}$-level scene hierarchy as follows:
    \begin{gather}
        \mathcal{J}^{\rm{distill}}_{i}=\frac{1}{|\mathcal{V}|} sqrt[ \sum_{q_i \in\ \mathcal{L}_i} (score^{\rm{t}}_{q_i}-score^{\rm{nt}}_{q_i})^2], \\
        \mathcal{J}^{\rm{distill}}=\beta_{1}^{\rm{level}}\mathcal{J}_{1}^{\rm{distill}}+\beta_{2}^{\rm{level}}\mathcal{J}_{2}^{\rm{distill}},
    \end{gather}
    where $\beta_{1}^{\rm{level}}$ and $\beta_{2}^{\rm{level}}$ are hyper-parameters as coefficients of $\mathcal{J}_{1}^{\rm{distill}}$ and $\mathcal{J}_{2}^{\rm{distill}}$, and $\mathcal{L}_i$ denotes the set of the scene labels in the $i^{th}$-level hierarchical layer.

    In the end, the final objective function is defined as:
    \begin{gather}
        \mathcal{J}=\beta^{\rm{t}}\mathcal{J}^{\rm{t}}+\beta^{\rm{nt}}\mathcal{J}^{\rm{nt}}+\beta^{\rm{distill}}\mathcal{J}^{\rm{distill}},
    \end{gather}
    where $\beta^{\rm{t}}$, $\beta^{\rm{nt}}$ and $\beta^{\rm{distill}}$ are hyper-parameters as coefficients of $\mathcal{J}^{\rm{t}}$, $\mathcal{J}^{\rm{nt}}$ and $\mathcal{J}^{\rm{distill}}$.

    Moreover, to balance the performance and efficiency of the model, inspired by previous work \cite{CORR_DKNN_2015, ICLR_UDPI_2016, CVPR_STGK_2020}, the two modules are both utilized to participate in the training, but only the temporal module is employed for reasoning.
    When reasoning, given videos and the associated textual descriptions as queries, the model will first calculate the scene label scores through the temporal module, then take out the scene labels with the Top-K scores or the scores that exceed a specific threshold, and return them to users.
\section{EXPERIMENTS}
In this section, we conduct experiments on a real-world dataset to evaluate the performance of our proposed method.
\subsection{Dataset}
    We evaluate our model on one of the largest available video scene datasets, which is called the Koubei dataset. The Koubei Dataset contains metadata from Koubei Platform\footnote{www.koubei.com}, 
    including 63,977 videos with associated textual descriptions,
    and manually annotated hierarchical scene labels,
    where each video can correspond to multiple scene labels. 
    Besides, the scene label hierarchy has 6 $1^{st}$-level labels and 320 $2^{rd}$-level labels.

    We randomly split the Koubei dataset into training, validation, and testing sets. 
    Specifically, we randomly sample 1,000 videos into the validation and 1,000 videos into the testing sets, respectively, 
    and the remaining 61,977 videos are utilized as the training set.
\subsection{Evaluation Protocol and Parameters Settings}
    
    We evaluate the performance of different models using F1 score and RP@90\% as the evaluation metrics. 
    RP@90\% denotes the proportion of labelled videos when we add threshold constraints to make Accuracy reaches 90\%, 
    and this metric is widely employed in commercial products.

    In our experiments, we set the number of the sampled keyframes $N_f$ as 12, 
    the number of consecutive frames for 3D features $N_c$ as 16, 
    and the maximum number of extracted keywords $N_r$ as 10. 
    For simplicity, We set all the hyper-parameters $\beta^{\rm{t}}$, $\beta^{\rm{nt}}$, $\beta^{\rm{distill}}$, $\beta^{\rm{distill}}_{1}$, $\beta^{\rm{distill}}_{2}$, $\beta^{\rm{level}}_{1}$, $\beta^{\rm{level}}_{2}$ to the same value 1.
    We introduce \emph{ConceptNet} as the external knowledge graph, and the pretrained entity embeddings are provided by \emph{ConceptNet Numberbatch}~\cite{AAAI_CONCEPTNET_2017}.
    We utilize GeLU
    as the activation function, and employ \emph{dropout} with 50\% keep probability, 
    weight decay,
    and early stopping to alleviate overfitting. 
    To train our proposed model, we randomly initialize model parameters with a Gaussian distribution, and utilize AdamW \cite{ICLR_ADAMW_2019} algorithm for optimization. We further restrict the dimensions of the final representation vector of each video to be the same for fair comparisons.  
    We have tried different parameter settings, including the batch size of \{8, 16, 32\}, the latent feature dimension of \{192, 384, 768\}, the learning rate of \{1e-1, 3e-4, 1e-4, 3e-5, 1e-5\}. As the findings are consistent across the dimensions of latent vectors, if not specified, we only report the results based on the dimension of 768, which gives relatively good performance.

\subsection{Baselines}
    To evaluate the effectiveness of our model, we compare our proposed method with several state-of-the-art baselines.
    Specifically, the original task of the first four models is not video scene recognition, we adjust these models and perform the task of scene prediction based on the obtained video representations using them. The latter two models which are originally designed for scene recognition are directly utilized in experiments. 
    \begin{itemize}
        \item MMBT \cite{NEURIPS_MMBT_2019}: 
        introduces a supervised multi-modal biTransformer that jointly finetunes uni-modally pretrained text and image encoders. Besides, this model is the backbone of the temporal module of our model.
        \item LSCTA \cite{ACMMM_LSCTA_2020}: 
        employs a Transformer model as the backbone, and develops a framework to align sequences in different modalities to capture information.
        \item HGLTM \cite{ACMMM_HGLTM_2019}: 
        proposes a hierarchical model which can construct the temporal structure at frame-level and object-level successively, and extract pivotal information effectively from global to local, which improves the model capacity.
        \item STGK \cite{CVPR_STGK_2020}: 
        proposes a novel spatio-temporal graph network to explicitly exploit the spatio-temporal object interaction, which is crucial for video understanding.
        \item HCMFL \cite{TRANS_HCMFL_2019}: 
        develops a Hierarchy-dependent Cross-platform Multi-view Feature Learning framework, which jointly utilizes multi-platform data, object-scene deep features,
        and the hierarchical structure prior for category prediction from videos.
        \item HETFN \cite{TRANS_HETFN_2021}: 
        proposes a Hybrid-Attention Enhanced Two-Stream Fusion Network for the video recognition task, and develops a novel Global-Local Attention Module, which can be inserted into neural networks to generate enhanced visual features from video content. 
        \item $\text{Ours-S2}$: 
        This variant removes the Temporal Feature Learning module (S1) from the origin proposed model, 
        and keeps the Knowledge-enhanced Non-temporal Feature Learning module (S2) to learn the representation of videos. 
        \item $\text{Ours-S2}^\text{w/o-hier}$: 
        This variant additionally removes the loss functions for the $1^{st}$-level scene hierarchical layer on the basis of
        $\text{Ours-S2}$.
        \item $\text{Ours-S2}^\text{w/o-know\&w-temp}$: 
        This variant replaces the frame-level local feature fusion and the knowledge-enhanced video feature fusion in the non-temporal module by \emph{Transformer} on the basis of $\text{Ours-S2}$.
        \item $\text{Ours-S2}^\text{w/o-know}$: 
        This variant drops the proposed knowledge-enhanced video feature fusion module, 
        and utilizes the mean feature of the fused frame-level local features as the video-level non-temporal feature on the basis of $\text{Ours-S2}$.
        \item $\text{Ours-S2}^\text{w/o-kw}$: 
        This variant replaces the employed pretrained keyword entity embeddings by an equal number of randomly initialized embeddings shared by all videos on the basis of $\text{Ours-S2}$.
    \end{itemize}
\begin{table}[!t]
	\caption
		{
        Performance comparison among our model and all fully-trained baselines using the metrics RP@90\% and F1 score.
		}
	\label{experiment_result}
	\centering
	\begin{tabular}{l|cc}
    
    \specialrule{0.12em}{0pt}{0pt}
    \multirow{1}{*}{Model} &\small{RP@90\%}&\small{F1 score}\\
    \specialrule{0.12em}{0pt}{0pt}
	MMBT			&0.521&0.720 \\
	LSCTA			&0.324&0.706 \\
    HGLTM			&0.434&0.719 \\
    STGK			&0.452&0.707 \\
    HCMFL			&0.445&\textbf{0.735} \\
    HETFN			&0.483&0.710 \\
    \hline
    ${\text{Ours-S2}}$	&0.536&0.726 \\
    ${\text{Ours-S2}}^{\text{w/o-hier}}$	&0.511&0.731 \\
	${\text{Ours-S2}}^{\text{w/o-know\&w-temp}}$	&0.511&0.709 \\
	${\text{Ours-S2}}^{\text{w/o-know}}$		&0.504&0.707 \\
	${\text{Ours-S2}}^{\text{w/o-kw}}$		&0.531&0.706 \\
	OurModel		&\textbf{0.561}&\textbf{0.735} \\
    \specialrule{0.12em}{0pt}{0pt}  
  	\end{tabular}\vspace{-5pt} 
\end{table}
\subsection{Performances, Quantitative Analysis and Ablation Study}
    We evaluate all fully-trained models using the metrics F1 score and RP@90\%, and report the results 
    in Table~\ref{experiment_result}. 
    We have the following observations with respect to our experimental results.

    First, our proposed method achieves the best performance on Koubei Dataset using the metric RP@90\% and F1 score, demonstrating the effectiveness of our model.

    Second, compared with MMBT which is the backbone of the temporal module of OurModel, and models videos only from the temporal perspective,
    the whole OurModel achieves a 7.7\% performance gain on RP@90\% and 2.1\% performance gain on F1 score, 
    demonstrating the effectiveness of our proposed knowledge-enhanced non-temporal module. 
    Besides, OurModel achieves better performance than separate stream modules MMBT and $\text{Ours-S2}$, verifying that the two perspectives (\ie{the temporal and the knowledge-enhanced non-temporal perspectives}) are both valuable for the task of scene recognition and complementary to each other.

    Third, compared with $\text{Ours-S2}^\text{w/o-hier}$, $\text{Ours-S2}$ receives better performance, verifying that calculating loss function simultaneously for multi-level scene hierarchy can improve the performance of our model.

    Moreover, $\text{Ours-S2}$ obtain better performance than $\text{Ours-S2}^\text{w/o-know\&w-temp}$ and $\text{Ours-S2}^\text{w/o-know}$, demonstrating the effectiveness of the proposed frame-level local feature fusion and knowledge-enhanced video feature fusion.
    
    Finally, compared with $\text{Ours-S2}^\text{w/o-kw}$, 
    $\text{Ours-S2}$ achieves better performance, verifying that the introduced external knowledge is valuable to our proposed model.

\section{Conclusions and Future Work}
  With the explosive growth of video data in real-world applications, a comprehensive representation of videos becomes increasingly important. 
  In this paper, we address the problem of video scene recognition, whose goal is to learn a high-level video representation to classify scenes in videos. In this paper, we propose a novel two-stream framework to model video representations from the temporal and knowledge-enhanced non-temporal perspectives, 
  and integrate these two perspectives in an end-to-end manner by self-distillation. 
  Besides, we design a knowledge-enhanced feature fusion and label prediction method that contributes to naturally introducing knowledge into the task of video scene recognition. 
  We evaluated our model for scene recognition on a real-world dataset,
  and the experimental results demonstrate the effectiveness of our proposed model. 
  In addition, we also conducted ablation studies, which demonstrated the effectiveness of our proposed temporal and non-temporal two-stream framework and knowledge-enhanced feature fusion method, respectively. 
  In future work, we may pay more attention to the utilization of knowledge graphs, and try to leverage edge information in knowledge graphs to assist reasoning to enhance video understanding.

{\small
\bibliographystyle{ieee_fullname}
\bibliography{reference}
}

\end{document}